\documentclass{article} 
\usepackage{nips13submit_e,times}
\usepackage{hyperref}
\usepackage{url}
\usepackage{graphicx}

\newcommand{\bfx}{\mathbf{x}}

\newcommand{\bfw}{\mathbf{w}}
\newcommand{\Xcal}{\mathcal{X}}

\newcommand{\Fcal}{\mathcal{F}}

\newcommand{\Scal}{\mathcal{S}}
\newcommand{\E}{\mathbb{E}}

\usepackage{graphicx} 
\usepackage{subfigure} 

\usepackage{amssymb}
\usepackage{amsmath}

\usepackage{algorithm}
\usepackage{algorithmic}

\newtheorem{theorem}{Theorem}[section]
\newtheorem{proposition}[theorem]{Proposition}
\newtheorem{definition}[theorem]{Definition}

\newcommand{\RR}{\mathbb{R}}

\title{Principled Non-Linear Feature Selection }

\author{
Dimitrios Athanasakis\thanks{Department of Computer Science, University College London, London, UK}\\
\texttt{dathanasakis@cs.ucl.ac.uk} \\
\And
John Shawe-Taylor* \\
\texttt{jst@cs.ucl.ac.uk} \\
\AND
Delmiro Fernandez-Reyes\thanks{National Institute For Medical Research, London, UK} \\
\texttt{dfernan@nimr.mrc.ac.uk} \\
}

\nipsfinalcopy 

\begin{document}

\maketitle

\begin{abstract}

Recent non-linear feature selection approaches employing greedy optimisation of Centred Kernel Target Alignment (KTA) exhibit strong results in terms of generalisation accuracy and sparsity. However, they are computationally prohibitive for large datasets. We propose randSel, a randomised feature selection algorithm, with attractive scaling properties. Our theoretical analysis of randSel provides probabilistic guarantees for correct identification of relevant features under reasonable assumptions. RandSel's characteristics make it an ideal candidate for identifying informative learned representations. We've conducted experimentation to establish the performance of this approach, and present encouraging results, including a 3rd position result in the recent ICML black box learning challenge as well as competitive results for signal peptide prediction, an important problem in bioinformatics. 
\end{abstract}


\section{Introduction}

Feature selection is an important aspect in the implementation of machine learning methods. The selection of informative features can reduce generalisation error as well as storage and processing requirements for large datasets. In addition,  parsimonious models provide valuable insight into the relations underlying elements of the process under examination. There is a wealth of literature on the subject of feature selection when the relationship between variables is linear. Unfortunately when the relation is non-linear feature selection becomes substantially more nuanced. 

Kernel methods excel in modelling non-linear relations. Unsurprisingly, a number of kernel-based feature selection algorithms have been proposed.  Early propositions, such as Recursive Feature Elimination (RFE) [1] can be computationally prohibitive, while attempts to learn a convex combination of low-rank kernels may fail to encapsulate nonlinearities in the underlying relation. Recent approaches using explicit kernel approximations can capture non-linear relations, but increase the storage and computational requirements. 

\subsection{Related Work}

Our approach makes extensive use of \emph{Kernel Target Alignment} (KTA) [2,3], as the empirical estimator for the Hilbert-Schmidt Independence Criterion (HSIC). Work on HSIC [4] provides the foundation of using the alignment of centred kernel matrices as the basis for measuring statistical dependence. The Hilbert-Schmidt Independence criterion is the basis for further work in  [5], where greedy optimisation of centred alignment is employed for feature selection. Additionally, [5] identifies numerous connections with other existing feature selection algorithms which can be considered as instances of the framework.

Stability selection [6] is a general framework for variable selection and structure estimation of high dimensional data. The core principle of stability selection is to combine subsampling with a sparse variable selection algorithm.  By repeated estimation over a number of different subsamples, the framework keeps track of the number of times each variable was used, thus maintaining an estimate for the importance of each feature. In this work, we propose a synthesis of the two aforementioned approaches through a randomised feature selection algorithm based on estimating the statistical dependence between bootstrapped random subspaces of the dataset in RKHS. The dependence estimation of random subsets of variables is similar to the approach of [11], which is extended through bootstrapping and carefully controlled feature set sizes.  

Our proposal is simple to implement and compares favourably with other methods in terms of scalability. The rest of the paper is structured as follows: \emph{Section 2} presents the necessary  background  on feature selection for kernel-based learning. \emph{Section 3} introduces a basic randomised algorithm for nonlinear feature selection, along with some simple examples, while \emph{Section 4} provides some analysis.\emph{Section 5} provides examples of how randSel can be effectively utilised as an important constituent of representation learning. \emph{Section 6} provides experimental results, and the brief discussion of \emph{Section 7} concludes this paper. 

\section{Preliminaries}

We consider the supervised learning problem of modelling the relationship between a $m \times n$ input matrix $X$ and a corresponding $m \times n'$ output matrix $Y$. The simplest instance of such a problem is binary classification where the objective is the learning problem is to learn a function $f: \textbf{x} \rightarrow \textbf{y}$ mapping input vectors $\textbf{x}$ to the desired outputs $\textbf{y}$.
In the binary case we are presented with a $m \times n $ matrix $X$ and a vector of outputs $\textbf{y}$,  $y_i \in \{+1,-1 \}$
Limiting the class of discrimination functions to linear classifiers we wish to find a classifier 
$$f(\textbf{x}) = \sum_i w_i x_i = \langle \textbf{w, x}\rangle$$

The linear learning formulation can be generalised to the nonlinear setting through the use of a  nonlinear feature map $\phi(x)$, leading to the kernelized formulation: 
$$f(x)=\langle w, \phi(x)\rangle = \langle \sum_i a_i y_i \phi(x_i), \phi(x)\rangle = \sum_i a_i y_i k(x_i,x) $$

The key quantit? of interest in our approach is the centred kernel target alignment which is the empirical estimator of the HSIC[4], which measures statistical dependence in RKHS:  
$$a(C_x,C_y)= { {\langle C_x,C_y \rangle_F}\over{ \|C_x\|_F \|C_y\|_F}} = {{\sum_{i,j}c_{x_{ij}} c_{y_{ij}}}\over { \sum_{i,j}\|c_{x_{ij}}\|\sum_{i,j}\|k_{y_{ij}}\| }  }$$

The matrices $C_x$ and $C_y$ correspond to centred kernels on the features $X$ and outputs $Y$ and are computed as:

$$C=\left[ I-{{11^T}\over m} \right] K \left[ I-{{11^T}\over m} \right]  $$

where $1$, in the above equation denotes the m-dimensional vector with all entries set equal to one.

\section{Development of key ideas}

The approach we will take will be based on the following well-known observation that links kernel target alignment with the degree to which an input space contains a linear projection that correlates with the target.

\begin{proposition}
Let $P$ be a probability distribution on the product space $\Xcal \times \RR$, where $\Xcal$ has a projection $\phi$ into a Hilbert space $\Fcal$ defined by a kernel $\kappa$. 
We have that
\begin{eqnarray*}
\sqrt{\E_{(\bfx,y) \sim P, (\bfx', y')\sim P}[y y' \kappa(\bfx,\bfx')]} =&& \\
&& \hspace{-5cm}= \sup_{\bfw: \|\bfw\| \leq 1} \E_{(\bfx,y)\sim P}[y\langle\bfw,\phi(\bfx)\rangle]
\end{eqnarray*}
\end{proposition}
{\bf Proof:}
\begin{eqnarray*}
\sup_{\bfw: \|\bfw\| \leq 1} \E_{(\bfx,y)\sim P}[y\langle\bfw,\phi(\bfx)\rangle ] =&& \\
&&\hspace*{-5cm}= \sup_{\bfw: \|\bfw\| \leq 1} \left\langle\bfw,\E_{(\bfx,y)\sim P}[\phi(\bfx) y]\right\rangle \\
&&\hspace*{-5cm}= \left\| \E_{(\bfx,y)\sim P}[\phi(\bfx) y]\right\|\\
&&\hspace*{-5cm}=\sqrt{ \int \int dP(\bfx,y)dP(\bfx',y') \langle \phi(\bfx),\phi(\bfx')\rangle 
yy'}\\
&&\hspace*{-5cm}= \sqrt{\E_{(\bfx,y) \sim P, (\bfx', y')\sim P}[y y' \kappa(\bfx,\bfx')] }
\end{eqnarray*}

The proposition suggests that we can detect useful representations by measuring kernel target alignment. For non-linear functions the difficulty is to identify which combination of features creates a useful representation. We tackle this problem by sampling subsets $S$ of features and assessing whether on average the presence of a particular feature $i$ contributes to an increase $c_i$ in the average kernel target alignment. In this way we derive an empirical estimate of a quantity we will term the contribution.

\begin{definition}
The {\em contribution} $c_i$ of feature $i$ is defined as
\[
c_i = \E_{S \sim \Scal_i}\left[\E_{(\bfx,y) \sim P, (\bfx', y')\sim P}[y y' \kappa_S(\bfx,\bfx')]\right] - 
\E_{S' \sim \Scal_{\setminus i}}\left[\E_{(\bfx,y) \sim P, (\bfx', y')\sim P}[y y' \kappa_{S'}(\bfx,\bfx')]\right],
\]
where $\kappa_S$ denotes the (non-linear) kernel using features in the set $S$ (in our case this will be a Gaussian kernel with equal width), $\Scal_i$ the uniform distribution over sets of features of size $\lfloor n/2\rfloor + 1$ that include the feature $i$, $\Scal_{\setminus i}$ the uniform distribution over sets of features of size $\lfloor n/2 \rfloor$ that do not contain the feature $i$, and $n$ is the number of features. 
\end{definition}

Note that the two distributions over features $\Scal_i$ and $\Scal_{\setminus i}$ are matched in the sense that for each $S$ with non-zero probability in $\Scal_{\setminus i}$, $S \cup \{i\}$ has equal probability in $\Scal_i$. This approach is a straightforward extension of the idea of BaHsic [5].

We will show that for variables that are independent of the target this contribution will be negative. On the other hand, provided there are combinations of variables including the given variable that can generate significant correlations then the contribution of the variable will be positive.

\begin{definition}
We will define an {\em irrelevant} feature to be one whose value is statistically independent of the label and of the other features. 
\end{definition}

We would like an assurance that irrelevant features do not increase alignment. This is guaranteed for the Gaussian kernel by the following result.

\begin{proposition}
Let $P$ be a probability distribution on the product space $\Xcal \times \RR$, where $\Xcal$ has a projection $\phi_Si$ into a Hilbert space $\Fcal$ defined by the Gaussian kernel $\kappa_S$ on a set of features $S$. Suppose a feature $i \not \in S$ is irrelevant.
We have that
\[
\E_{(\bfx,y) \sim P, (\bfx', y')\sim P}[y y' \kappa_{S\cup\{i\}}(\bfx,\bfx')] \leq
\E_{(\bfx,y) \sim P, (\bfx', y')\sim P}[y y' \kappa_{S}(\bfx,\bfx')] 
\]
\end{proposition}
{\bf Proof (sketch):} Since the feature is independent of the target and the other features, functions of these features are also independent. Hence,
\begin{eqnarray*}
\E_{(\bfx,y) \sim P, (\bfx', y')\sim P}[y y' \kappa_{S\cup\{i\}}(\bfx,\bfx')]
&&= \E_{(\bfx,y) \sim P, (\bfx', y')\sim P}[y y' \kappa_{S}(\bfx,\bfx')\exp(- \gamma (x_i - x'_i)^2)]\\
&&\hspace*{-2.5cm}= \E_{(\bfx,y) \sim P, (\bfx', y')\sim P}[y y' \kappa_{S}(\bfx,\bfx')]\E_{(\bfx,y) \sim P, (\bfx', y')\sim P}[\exp(- \gamma (x_i - x'_i)^2)]\\
&&\hspace*{-2.5cm}= \E_{(\bfx,y) \sim P, (\bfx', y')\sim P}[y y' \kappa_{S}(\bfx,\bfx')]\alpha
\end{eqnarray*}
for $\alpha = \E_{(\bfx,y) \sim P, (\bfx', y')\sim P}[\exp(- \gamma (x_i - x'_i)^2)] \leq 1$.

\begin{figure}[h]
\begin{center}
\includegraphics[width=\textwidth]{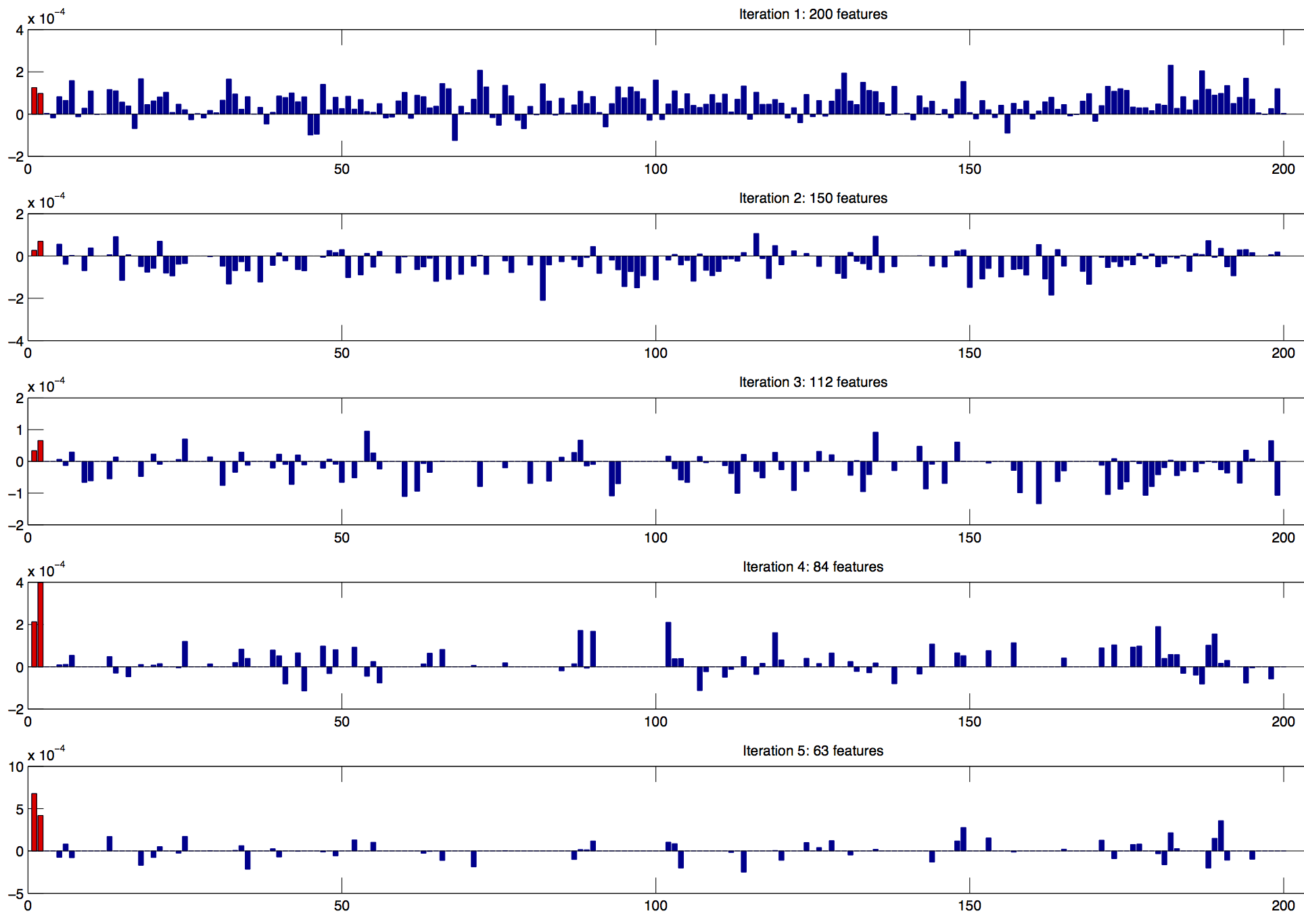}%
\end{center}
\caption{200-dimensional XOR classification problem using a subsample size of 100 and repeating over 1,000 random partitions of the variables. The expected contribution of the two relevant features is in red. Owing to the subsample size and the number of iterations, it is necessary to iteratively reject low contributing features.  It can also be seen that as more of the irrelevant features are removed in later iterations of the method, the expected contribution of the two relevant variables rises substantially.}
\label{xor}
\end{figure}

In fact the quantity $\alpha$ is typically less than 1 so that adding irrelevant features decreases the alignment. Our approach will be to progressively remove sets of features that are deemed to be irrelevant, hence increasing the alignment together with the signal to noise ratio for the relevant features. Figure \ref{xor} shows how progressively removing features from a learning problem whose output is the XOR function of the first two features both increases the alignment contributions and helps to highlight the two relevant features.
\begin{figure}[h]
\begin{center}
\includegraphics[width=\textwidth]{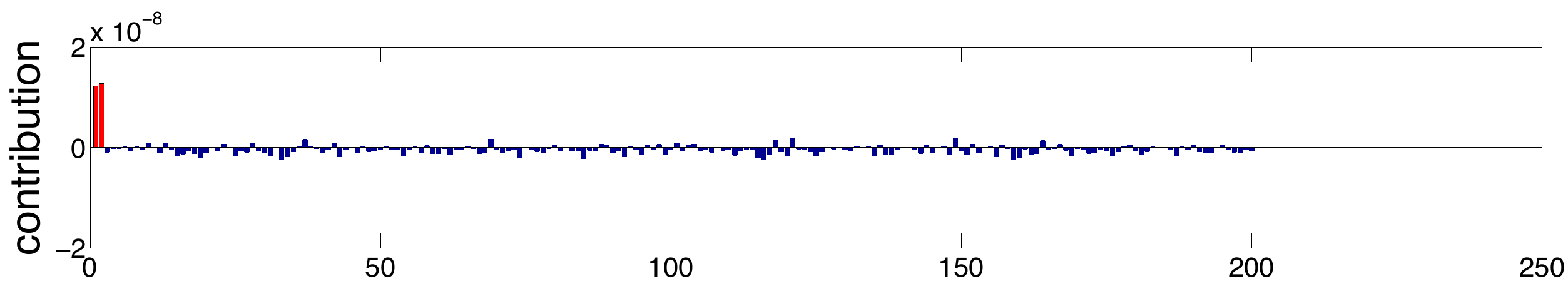}%
\end{center}
\caption{200-dimensional XOR classification problem, with a subsample size of 1,000 and repeated over 10,000 random partitions of the features. The expected contribution of the $\eta$-influential features, shown in red, are clearly separated from that of all irrelevant variables.}
\label{xorBig}
\end{figure}

We now introduce our definition of a relevant feature.
\begin{definition}
A feature $i$ will be termed $\eta$-{\em influential} when its contribution $c_i \geq \eta > 0$.
\end{definition} 

So far we have only considered expected alignment.  In practice we must estimate this expectation from a finite sample.  We will omit this part of the analysis as it is a straigthforward application of U-statistics that ensures that with high probability for a sufficiently large sample from $\Scal_i$ and $\Scal_{\setminus i}$ and of samples from $P$ (whose sizes depend on $\eta$, $\delta$, the number $k$ of $\eta$-influential variables and the number $T$ of iterations) an empirical estimate of the contribution of an $\eta$-influential variable will with probability at least $1 - \delta$ be greater than 0 for all of the fixed number $T$ of iterations of the algorithm.

Our final ingredient is a method of removing irrelevant features that we will term culling. At each iteration of the algorithm the contributions of all of the features are estimated using the required sample size and the contributions are sorted. We then remove the bottom 12.5\% of the features in this ordering. Our main result assures us that culling will work under the assumption that the irrelevant variables are independent. 

\begin{theorem}
\label{otherTheorem}
Fix $\eta> 0$. Suppose that there are $k$ $\eta$-influential variables and all other variables are irrelevant. Fix $\delta > 0$ and number $T$ of iterations. Given sufficiently many samples as described above the culling algorithm will with probability at least $1 - \delta$ remove only irrelevant variables.
\end{theorem}

{\bf Proof (sketch):} Through use of Hoeffding's inequality for U-statistics we can bound the deviation from the true expectation for all irrelevant variables $i$, $\hat{c}_i \leq {\eta \over 2} $ with probability at least $1-\delta$. Conversely, for all relevant variables will be within $\hat{c}_j\geq {\eta\over 2}$, with probability at least $1-\delta$. Therefore, provided sufficiently many samples, removing all variables with contribution estimated contribution $\hat{c}_i < \eta/2 $ will only remove irrelevant variables, and preserve all relevant variables. 

Figure \ref{xorBig} illustrates how for a large enough sample size and number of random partitions, the algorithm can identify the relevant variables with a high relative margin in the expected contribution. However, the sample size required to achieve the probabilistic guarantees of Theorem \ref{otherTheorem}, is too large for most practical settings. For this reason in our experimentation, we proceed to iteratively cull a smaller percentage of the bottom-contributing features at the end of each iteration. For example, the experiments here were performed with culling 12.5\% of the features after the end of each iteration, a process illustrated in figure \ref{xor}.

\section{Properties of the algorithm}

We now define our algorithm for randomised selection (randSel). Given a $m \times n$ input matrix $X$ and corresponding output matrix $Y$, randSel proceeds by estimating the individual contribution of features by estimating the alignment of a number of random subsamples that include $n\over 2$ and ${n\over 2}+1$ randomly selected features. This leads to an estimate for the expected alignment contribution of including a feature. The algorithm is parametrized by the number of subsamples $N$, a subsample size$n_b$ and a percentage $z\%$ of features that are dropped after $N$ subsamples. The algorithm proceeds iteratively until only two features remain. 

There are a number of benefits to this approach, aside from the tangible probabilistic guarantees. RandSel scales gracefully. Considering the computation of a kernel $k(x,x')$ for samples $x, x'$ atomic, the number of kernel computations for a single iteration are $ n_b^2 N$, which for a sensible choice of $N$ can be substantially smaller than the $m^2 n$ complexity of HSIC variants. For example setting $n_b= \sqrt{m}$ and $N=n$ an iteration would require $mn$ kernel element computations, and in addition this process is trivial to parallelize. 
\newcommand{\mean}{\mathrm{mean}}
\begin{algorithm}[h!]
   \caption{randSel}
   \label{alg:example}
\begin{algorithmic}
   \STATE {\bfseries Input:} input data $X$, labels $Y$, number of iterations $r$ subsample size $s$, number of features $n$, drop percentile proportion $z$, top percentile proportion $a$, number of occasions $t$
   \REPEAT   
   \FOR{$i=1$ {\bfseries to} $r$}
   \STATE $(X_{i},Y_{i})$ = Random subsample of size $s$ over ${n\over 2}$ randomly selected variables
   \STATE $a_i$= alignment($X_i, Y_i$)
   \STATE $(X^{(+)}_{i},Y^{(+)}_{i})$ = Random subsample of size $s$ over ${n\over 2}+1$ randomly selected variables
   \STATE $a_i^{(+)}$= alignment($X_i^{(+)}, Y_i^{(+)}$)
   \ENDFOR
   \FOR{$j=1$ {\bfseries to} $n$}   
   \STATE mean contribution $c_j = \mean_{i: j \in X^{(+)}_i} (a^{(+)}_i) - \mean_{i: j \notin X_i}(a_i)$,
   \ENDFOR
   \STATE drop the  $z$\% bottom-contributing features
   \IF{ fixing features}
   \IF{ $j$ top-contributor for $t$ consecutive times}
   	\STATE fix feature $j$
     \ENDIF
    \ENDIF 
   \UNTIL{no features left to fix, or only 2 features remain}
   \STATE Return Sequence of estimated contributions and Fixed Variables 
\end{algorithmic}
\end{algorithm}

\section{Feature Selection for learned representations}

Unsupervised feature learning algorithms such as sparse filtering [9] are often used to learn over-complete representations of data. The depth of a learning architecture refers to the composition of different levels of non-linear operations in the learned function. This suggests that employing feature selection to refine a set of learned representations, would substantially benefit from capturing non-linear interactions between the learned features. 
Utilising randSel for feature selection in this setting is predicated on a number of properties.  RandSel is readily applicable to a large sample size, a key property for the large sample sizes typically involved in representation learning. In addition, the algorithm is readily applicable to domains that have some structure. The multi-class structure of the black box challenge is an example where this property is important. This is a shared property of all the HSIC-variants. Finally, randSel is granular. Here, granularity refers to the fact that at the end of each iteration the algorithm returns a list of the remaining features and their expected contributions, which leads to a series of kernels of increased granularity. This can be a highly attractive property when using MKL for the final prediction. 

\subsection{Prediction}

RandSel produces progressively fine-grained combinations of features. A prediction mechanism effectively utilising the increasingly granular combinations of features comprises the last step of our approach, where we take a boosting approach based on LPBOOST-MKL [10]. The architecture  proceeds by building a number of Gaussian kernels, parametrised by the different sets of features they are defined on, and a kernel bandwidth parameter $\sigma$ as 
$$ \kappa_{(s_i,\sigma)}(x,x') = \exp(-\sigma ( x^{(s_i)}-x'^{(s_i)} )^{2}  $$,
Where, $x^{(s_i)}, x'^{(s_i)}$ are vectors containing only variables included in the set $s_i$. For simplicity assume there are $n_K$ distinct combinations of feature sets $s_i$ and corresponding bandwidths $\sigma$. We define a kernel on each such combination of features and bandwidth. Kernel ridge regression was used to generate the individual weak predictors in our architecture. Thus, each weak predictor has the form 
$$h(x,x')= \sum_i a_i \kappa_{s_j,\sigma}(x_i,x')$$ 

The algorithm then computes the classification rule, which is a convex combination of the weak learners, through the following linear program:

\begin{equation*}
\begin{aligned}
& {\text{minimize }} \beta \\
 \text{s.t. } &\sum_{i=1}^m u_i y_i H_ij \leq \beta\\
 &\sum_{i=1}^m u_i = 1\\
 & 0 \leq u_i, \leq D
\end{aligned}
\end{equation*}

Where $D$ is a regularization parameter. Provided a sensible range of kernel bandwidths $\sigma$ is specified, the final LPBoost classifier only requires tuning the regularization parameter $D$. In our search for simplicity, this is a tangible benefit, substantially reducing the search space of parameter combinations, to tuning this single regularization parameter. 
\section{Results} 

\subsection{Results on the ICML Black Box Learning Challenge}

We used our proposal in the recent ICML 2013 Challenges in Representation Learning Black Box Learning challenge [8][12]. The dataset used in the challenge was an obfuscated subset of the Street View House Numbers dataset \cite{SVHN}. The original data were projected down to 1875 dimensions by multiplication with a random matrix, and the organizers did not reveal the source of the dataset was SVHN until the competition was over. The training set comprises only 1,000 labelled samples, while an additional 130,000 samples were provided for the purposes of unsupervised pre-training. 

For our submissions, cross validation was used to select the number of features to learn with sparse filtering, with our best solution using a set of 625 learned features. Randsel was then used to select combinations of the 625 learned features dropping 12.5\% of the least contributing features at the end of each iteration. The resulting set of 34 different sets of features was combined with 75 different $\sigma$ parameters to result in 2550 weak learners. The regularisation parameter $D$ was also set through cross validation. This approach led to a generalisation accuracy of 68.44\% on the public and 68.48\% on the private leaderboards, ranking third in both cases out of a total of 218 teams.

\subsection{Application to cleavage site prediction}

Signal peptides are amino-acid sequences found in transported proteins that selectively guide the distribution of the proteins to specific cellular compartments. Often referred to as the zip-code sequences, owing to their role in sub-cellular localization, a substantial body of work is devoted to predicting the cleavage site of signal sequences. Current literature establishes the importance of a number of physicochemical properties of the signal sequence in determining the cleavage site location. The experimental pipeline presented in this section further supplements this approach, by learning a feature representation of multiple physicochemical property encodings. 

The Predisi dataset \cite{Predisi} of eukaryotic signal sequences was used for experimentation. Initial filtering produced a dataset of 2,705 unique signal peptide sequences, with a sequence length of 50 amino-acids. The approach used for cleavage site prediction breaks each individual sequence into smaller windows. Cross validation was used to estimate the parameters relating to the window size. The resulting convention was to use windows that contain 9 aminoacids prior to what we deem the target of the window and 2 aminoacids following that position. For an individual prediction to be considered accurate, the window predicted as most likely to contain the cleavage site in its target position, must coincide with the actual window containing the cleavage target site for the sequence.

With the window parameters chosen through cross validation, this results in each individual sequence of 50 aminoacids producing 39 windows with a length of 12 aminoacids each. The resulting process generates a dataset comprising of 105,495 windows. The entirety of 54 distinct physicochemical encodings offered by the Matlab bioinformatics toolbox was used for numerical representation of each sequence window, which is then represented by a 648-dimensional vector of physicochemical properties.At this point, sparse filtering learns an overcomplete representation comprising of 1500 learned features. This process generates a dataset comprising of 105,495 1500-dimensional samples. 

There are three interesting questions which the experiments where designed to address. Concretely, the experimental comparisons are designed to establish the performance gains from using a learned representation using sparse filtering, over learning in the original feature space, using randSel for feature selection as opposed to other possible feature selection methods, and finally establishing the importance of multiple kernel learning, used for prediction.   

To this end, a number of competing solutions were implemented. The shallow approach uses the raw physichochemical properties for prediction. For comparing the performance of different feature selection algorithms on learned representations, $\ell_1$-logistic regression stability selection and nonlinear SVM-RFE are used in addition to randSel. Finally we compare the performance of a prediction rule relying on a single gaussian kernel SVM, to the performance of MKL. The experiments employed libSVM [17] as an svm solver and SPAMS [16] for solving sparse logistic regression in the stability selection framework.

The large size of the dataset, as well as the fact that it is highly imbalanced make for some challenges. Stability selection can readily be applied to a problem of this dimensionality. While deterministic HSIC-variants are ill-equipped to deal with the size of the resulting kernel in most current commodity hardware, the use of sampling in randSel largely alleviates the problems related with size. In order to address the issue of the imbalance, subsamples where both classes are equally represented were used. 

In terms of producing the actual prediction the experiments examine two options. The first is using a chunking, Gaussian kernel SVM, which is the same approach that enables the use of RFE, which is used in combination with the various feature selection methods.  Using vanilla LPBoost-MKL for prediction is prohibitive, owing to the memory requirements of storing the kernel matrices. The approach to rectify this problem is to use subsampling from the negative class, combined with kernel ridge regression as a weak predictor in our boosting framework. This results in a mixed-norm MKL formulation which effectively addresses the limitation of not being able to store the kernel matrices. For the $l_2$ regularization parameter $\lambda$ of individual weak predictors, a very small range of parameters was used. The $l_1$ regularization parameter $D$ for the LPBoost prediction rule was set through cross-validation.   

\subsection{Experimental Results} 

Table \ref{peptideRez} summarizes the results for the different attempted approaches. Using the original feature representation with a non-linear SVM leads to a generalization accuracy of 67.2\%. This is substantially smaller than all the approaches that rely on the learned feature representation. This suggests that there are performance gains to be had in using a learned representation.

\begin{table}[h!tbp]
\caption{Accuracy for the different approaches discussed in section 6.2 when applied to the signal peptide problem.}
\label{peptideRez}
\begin{center}
\begin{tabular}{ l l l l l l l l l|| || }
\hline
\multicolumn{1}{|c}{\bf Method}  &\multicolumn{1}{c|}{\bf Accuracy (\%) }\\
\hline
Original Representation  & 67.20 $\pm$ 4.71 \\
Sparse Filtering + Stab. Sel. & 71.46 $\pm$ 2.93 \\
Sparse Filtering + RFE  &  71.81 $\pm$ 2.79  \\
Sparse Filtering + RandSel & 72.75 $\pm$ 2.85 \\
Sparse Filtering + RandSel \& MKL & 75.28 $\pm$ 1.91 \\
\end{tabular}
\end{center}
\end{table}

In terms of using feature selection on the learned representation, the results indicate that randSel has an edge over the two other methods, with RFE also performing slightly better than $l_1$-regularized logistic regression-based stability selection. The learned representation offers a case where it is reasonable to suspect benign non-linear collusion between features, something that both RFE and randSel are designed to take advantage of, and the large sample size allows for increased confidence when inferring such relationships. The fact that randSel outperforms RFE suggests that RFE's reliance on the support vectors for feature selection can negatively bias the feature selection procedure. Finally, the use of MKL for prediction further improves the results.  Direct comparisons to state-of-the art methods for cleavage prediction is difficult as the reported accuracy highly depends on the dataset and modelling assumptions, such as the original sequence length. Our proposed approach appears to outperform SignalP's  \cite{SignalP} reported accuracy of 72.9\% for eukaryotic sequences, but it must be noted that SignalP operates under different modelling assumptions and more extensive testing is necessary to account for that and to ascertain the significance of this finding. 
\section{Conclusions}

In this paper we propose randSel, a new algorithm for non-linear feature selection based on randomised estimates of HSIC. RandSel, stochastically estimates the expected importance of features at each iteration, proceeding to cull uninformative features at the end of each iteration. Our theoretical analysis gives strong guarantees for the expected performance of this procedure which is further demonstrated by testing on a number of real and artificial datasets. 

Additionally, we presented a simple system that produces a classification rule based on non-linear learned feature combinations of increasing granularity. The architecture of the system comprises a fast, unsupervised feature learning mechanism, randomised non-linear feature selection and a multiple kernel learning based classifier. The guiding principle of this approach is to use simple components that require minimal parameter tuning, with components further down the pipeline making up for the potential shortcomings upstream. Indeed, the three different constituents of this architecture, require minimal parameter tuning and scale gracefully, and the experimental results on both datasets we employed appear to validate the approach.

\newpage
\appendix Supplementary Material

\begin{figure}[!htb]
\begin{center}
\includegraphics[width=0.75\textwidth]{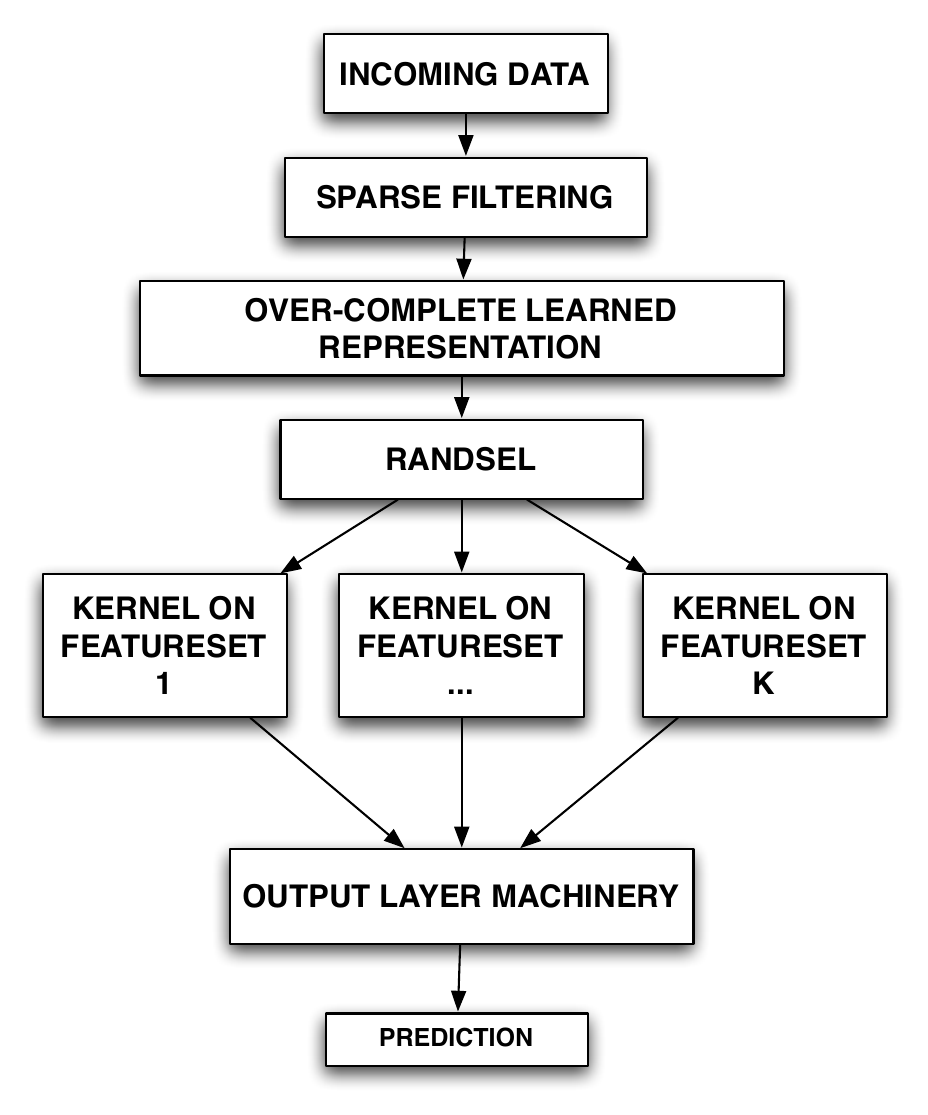}%
\end{center}
\caption{The architecture employed in our experiments; randSel is applied on the features learned by sparse filtering, producing a number of nonlinear combinations of learned features of increasing granularity. A number of kernels is defined on these nonlinear combinations of features, and multiple kernel learning is used for the overall prediction.}
\label{architecture}
\end{figure}

\begin{figure}[!hbt]
\begin{center}
\includegraphics[width=0.75\textwidth]{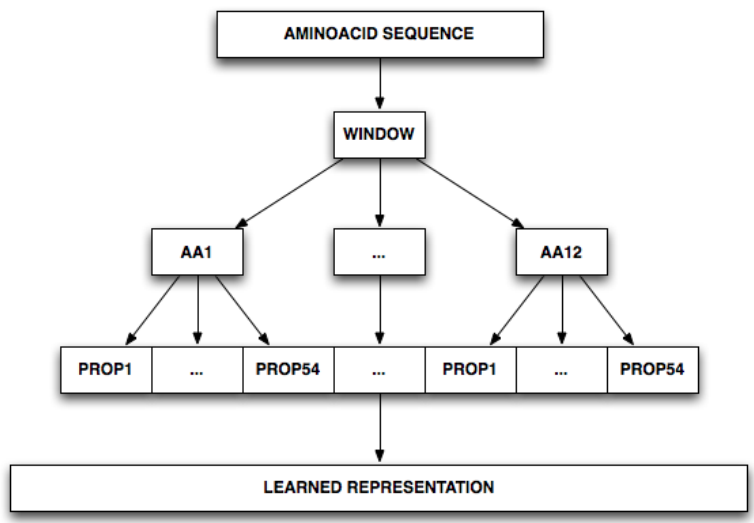}%
\end{center}
\caption{How the learned representation is for the signal peptide problem is generated. The amino-acid sequence is broken into smaller windows. Each amino-acid in the window is represented by its 54 distinct physicochemical properties. Sparse filtering is used to learn a representation for this encoding.}
\label{repGen}
\end{figure}

\begin{thebibliography}{14}

\small{

\bibitem{Guyon2002}
Guyon, Isabelle, Jason Weston, Stephen Barnhill, and Vladimir Vapnik (2002)
 Gene selection for cancer classification using support vector machines. 
 Machine learning 46, no. 1-3 : 389-422.

\bibitem{2}
Shawe-Taylor, N., and A. Kandola (2002) 
On kernel target alignment. 
Advances in neural information processing systems 14: 367.

\bibitem{3}
Cortes, Corinna, Mehryar Mohri, and Afshin Rostamizadeh (2012)
Algorithms for learning kernels based on centered alignment. 
The Journal of Machine Learning Research 13: 795-828.

\bibitem{4}
Gretton, Arthur, Olivier Bousquet, Alex Smola, and Bernhard Sch�lkopf (2005) 
Measuring statistical dependence with Hilbert-Schmidt norms. 
In Algorithmic learning theory, pp. 63-77. Springer Berlin Heidelberg.

\bibitem{5}
Song, Le, Alex Smola, Arthur Gretton, Justin Bedo, and Karsten Borgwardt (2012) 
Feature selection via dependence maximization. 
The Journal of Machine Learning Research 98888: 1393-1434.

\bibitem{6}
Meinshausen, Nicolai, and Peter B�hlmann (2012)
Stability selection. 
Journal of the Royal Statistical Society: Series B (Statistical Methodology) 72, no. 4: 417-473.

\bibitem{7}
Weston, Jason, Sayan Mukherjee, Olivier Chapelle, Massimiliano Pontil, Tomaso Poggio, and Vladimir Vapnik (2001) 
Feature selection for SVMs. 
Advances in neural information processing systems: 668-674.

\bibitem{8}
Challenges in Representation Learning: The Black Box Learning Challenge:
\url{https://www.kaggle.com/c/challenges-in-representation-learning-the-black-box-learning-challenge}

\bibitem{9}
Ngiam, Jiquan, Pang Wei Koh, Zhenghao Chen, Sonia Bhaskar, and Andrew Y. Ng.(2011) 
Sparse filtering.
Advances in Neural Information Processing Systems 24: 1125-1133.

\bibitem{10}
Tristan Fletcher, Zakria Hussain, John Shawe-Taylor (2010)
Multiple Kernel Learning on the Limit Order Book,
Proceedings of the First Workshop on Applications of Pattern Analysis 

\bibitem{11}
Somol, Petr, Jiri Grim, and Pavel Pudil. (2011) 
Fast dependency-aware feature selection in very-high-dimensional pattern recognition. 
In Systems, Man, and Cybernetics (SMC) 
2011 IEEE International Conference on, pp. 502-509. IEEE
}

\bibitem{12}
Goodfellow, I. J., Erhan, D., Carrier, P. L., Courville, A., Mirza, M., Hamner, B., ... \& Bengio, Y.  (2013)
Challenges in Representation Learning: A report on three machine learning contests. 
Proceedings of the 20th International Conference on Neural Information Processing 

\bibitem{Predisi}
Hiller, Karsten, Andreas Grote, Maurice Scheer, Richard MŸnch, and Dieter Jahn. "PrediSi: prediction of signal peptides and their cleavage positions." Nucleic acids research 32, no. suppl 2 (2004): W375-W379.

\bibitem{SignalP}
Petersen, Thomas Nordahl, S¿ren Brunak, Gunnar von Heijne, and Henrik Nielsen. "SignalP 4.0: discriminating signal peptides from transmembrane regions." Nature methods 8, no. 10 (2011): 785-786.

\bibitem{SVHN}
Yuval Netzer, Tao Wang, Adam Coates, Alessandro Bissacco, Bo Wu, Andrew Y. Ng Reading Digits in Natural Images with Unsupervised Feature Learning NIPS Workshop on Deep Learning and Unsupervised Feature Learning 2011

\bibitem{Spams}
Jenatton, Rodolphe, Julien Mairal, Francis R. Bach, and Guillaume R. Obozinski. "Proximal methods for sparse hierarchical dictionary learning." In Proceedings of the 27th International Conference on Machine Learning. 2010.

\bibitem{libsvm}
Chang, Chih-Chung, and Chih-Jen Lin. "LIBSVM: a library for support vector machines." ACM Transactions on Intelligent Systems and Technology (TIST) 2, no. 3 (2011): 27.

\end{thebibliography}
\end{document}